\documentclass[runningheads]{llncs}

\usepackage{algorithm}
\usepackage{algpseudocode}
\usepackage{pdflscape}
\usepackage[table]{xcolor}
\algnewcommand\algorithmicforeach{\textbf{for each}}
\algdef{S}[FOR]{ForEach}[1]{\algorithmicforeach\ #1\ \algorithmicdo}
\algnewcommand{\algorithmicand}{\textbf{ and }}
\algnewcommand{\AND}{\algorithmicand}

\usepackage{subcaption}
\usepackage{todonotes}
\usepackage{soul}

\newcommand{\review}[1]{\textcolor{black}{#1}}

\usepackage[T1]{fontenc}
\usepackage{graphicx}
\begin{document}
\title{Improving Sampling Methods for Fine-tuning SentenceBERT in Text Streams}
\titlerunning{Impr. Sampl. Methods for Fine-tuning SentenceBERT in Text Streams}
%
\author{Cristiano M. Garcia\inst{1,3}\orcidID{0000-0002-7475-146X} \and
Alessandro L. Koerich\inst{2}\orcidID{0000-0001-5879-7014} \and
Alceu de S. Britto Jr.\inst{3,4}\orcidID{0000-0002-3064-3563} \and Jean Paul Barddal\inst{3}\orcidID{0000-0001-9928-854X}}
\authorrunning{C. M. Garcia et al.}
%
\institute{Instituto Federal de Santa Catarina (IFSC), Câmpus Caçador, Brazil \email{cristiano.garcia@ifsc.edu.br} \and
École de Technologie Supérieure (ÉTS), Université du Québec, Montréal, Canada\\
\email{alessandro.koerich@etsmtl.ca}\\\and
Pontifícia Universidade Católica do Paraná (PUCPR), Curitiba, Brazil\\
\email{\{alceu,jean.barddal\}@ppgia.pucpr.br} \and
Universidade Estadual de Ponta Grossa (UEPG), Ponta Grossa, Brazil
}
\maketitle              
\begin{abstract}


The proliferation of textual data on the Internet presents a unique opportunity for institutions and companies to monitor public opinion about their services and products. Given the rapid generation of such data, the text stream mining setting, which handles sequentially arriving, potentially infinite text streams, is often more suitable than traditional batch learning. While pre-trained language models are commonly employed for their high-quality text vectorization capabilities in streaming contexts, they face challenges adapting to concept drift—the phenomenon where the data distribution changes over time, adversely affecting model performance. Addressing the issue of concept drift, this study explores the efficacy of seven text sampling methods designed to fine-tune language models, thereby mitigating performance degradation selectively. We precisely assess the impact of these methods on fine-tuning the SBERT model using four different loss functions. Our evaluation, focused on Macro F1-score and elapsed time, employs two text stream datasets and an incremental SVM classifier to benchmark performance. Our findings indicate that Softmax loss and Batch All Triplets loss are particularly effective for text stream classification, demonstrating that larger sample sizes correlate with improved macro F1 scores. Notably, our proposed WordPieceToken ratio sampling method significantly enhances performance with the identified loss functions, surpassing baseline results.

\end{abstract}

\section{Introduction}

The Internet has become part of daily life around the world. People and systems generate a vast amount of textual data through the Internet. Individuals can chat with others, review products and services, and share comments and opinions through social media platforms, frequently working as social sensors \cite{suprem2019assed}. Learning from social media posts can be relevant for institutions and governments, helping them quickly detect and respond to events \cite{garcia2023event,pohl2018batch}, for example.

Automatically learning from textual data leveraging machine learning mechanisms brings several challenges in batch processing, such as text standardization and vectorization. Text vectorization is essential since most machine learning methods expect numeric vectors as input. Traditional vector representations, such as Bag-of-Words (BOW) \cite{harris1954distributional} and Term Frequency - Inverse of Document Frequency (TF-IDF) \cite{salton1988term}, can generate very-high-dimensional vectors, which can be disadvantageous to the machine learning model, increasing the computational cost.

In a textual stream scenario, the challenges are augmented. Due to the stream characteristics, e.g., data arriving on an instance-basis or in small batches and resource limitations \cite{bifet2023machine,gama2014survey}, generating vector representations through BOW and TF-IDF is complex. For instance, if the vector representations are generated in the first batch, new words in subsequent batches will not be represented. On the other hand, generating the representations as the batches arrive can lead to variable-dimension representations \cite{garcia2023concept}, a challenge since most machine learning algorithms require a fixed-dimension input.

Therefore, pre-trained language models have become popular in batch and stream scenarios due to their time-saving characteristics \cite{garcia2023concept,thuma2023benchmarking}. SentenceBERT (SBERT) \cite{reimers2019sentencebert} is a popular pre-trained language model specific for sentence embedding generation. Although pre-trained models save time since training a language model from scratch is costly, adjustments may be necessary for domain adaptation. In addition, changes in data distribution over time (concept drift) are frequent phenomena in real-world data and can degrade a machine-learning model's performance \cite{gama2014survey}. In textual data streams, those changes can emerge from sentiment changes, the appearance of particular words in different contexts, and so on \cite{garcia2023concept}. Furthermore, computational linguistic studies using diachronic datasets attest that writing patterns change and word meanings evolve over time, e.g., semantic shift \cite{bravo2022incremental}. This work refers to changes as concept drift since semantic shifts are generally related to linguistic studies, which use long timespan, or diachronic, datasets and investigate those changes on a deeper, linguistic level. Therefore, to adapt to concept drift or a new domain, for example, it is important to adapt the language model. The fine-tuning process is a popular deep-learning-related process and can help adapt the language model \cite{lee2020biobert,schneider2023cardiobertpt}. 

Due to the computational cost of the fine-tuning process, selecting representative instances to fine-tune the language model may provide valuable information while reducing the time spent \cite{amba2021dynamic,sharir2020cost}. In this paper, we score the ability of different sampling methods in text selection for fine-tuning purposes. We also propose a sampling method, i.e., WordPieceToken ratio, whose results were promising in most scenarios evaluated. Considering the text stream setting, we assessed these methods intrinsically, i.e., in a downstream task. We also evaluated three versions of these sampling methods modified to account for the classes, totaling seven sampling methods.



The contributions of this paper are four-fold: (a) an extensive comparison among text sampling methods for fine-tuning purposes; (b) an analysis of the impact of the sampling methods considering the text stream setting; (c) an evaluation of loss functions for fine-tuning SBERT, and (d) a novel textual sampling method based on the ratio between Wordpieces and tokens of a text. The term Wordpieces represents a subword partition system present in BERT \cite{devlin2019bert} that allows handling out-of-vocabulary tokens.

This paper is organized as follows: Section \ref{sec:background} presents important concepts for understanding this paper, including text streams and SentenceBERT. Section \ref{sec:text-based-sampling-methods} presents the text-based sampling methods evaluated in this paper. Section \ref{sec:experimental-results} describes the experimental protocol, with datasets, settings, evaluation scenario, and results. Finally, Section \ref{sec:conclusion} concludes this paper.

\section{Background}
\label{sec:background}

This section presents core concepts for understanding this paper. We introduce text stream mining, SentenceBERT, and its fine-tuning process. 

\subsection{Text Stream Mining}

According to Bifet et al.~\cite{bifet2023machine}, data streams are ``an abstraction that allows real-time analytics''. In a data stream, the items arrive individually or in small sequential batches, and the stream itself can be infinite \cite{bifet2023machine,gama2014survey}. Different learning approaches have been developed to learn from data streams, including, for instance \cite{gama2014survey}, the ability to learn incrementally, single-pass operations, elimination of input data as soon as possible after learning from it, and consumption of modest resources, i.e., processing power and time.

Text streams are a specialization of data streams in which texts arrive over time \cite{garcia2023concept}. The challenges are extended in this scenario, mainly comprising natural language processing (NLP), such as text standardization, vocabulary, and representation maintenance. These NLP-related processing are challenging due to their complexity, which should meet the text stream constraints.

Frequently, text-related approaches leverage pre-trained language models \cite{garcia2023concept,thuma2023benchmarking}. Using pre-trained models can help save time since training a language model from scratch is computationally costly in time and resources \cite{sharir2020cost}. In addition, the language model can easily be reused in different scenarios. However, an important drawback is that texts generally suffer from concept drift. Concept drift is a phenomenon frequently observed in real-world datasets and corresponds to changes in data distribution over time \cite{gama2014survey,garcia2023concept}. Leveraging pre-trained language models without accounting for concept drift can decrease performance in the downstream task since the texts would be represented using relatively old representations \cite{garcia2023concept}.

This paper leverages a pre-trained language model and evaluates the use of fine-tuning for language model updates in text stream settings, a less costly approach than training from scratch. In this paper, the selected language model is SentenceBERT \cite{reimers2019sentencebert}.

\subsection{SentenceBERT}

SentenceBERT (SBERT) is an architecture that leverages pre-trained BERT models \cite{reimers2019sentencebert}, such as BERT \cite{devlin2019bert} and RoBERTa \cite{liu2019roberta} models. SBERT leverages siamese networks to generate semantically meaningful representations that are compared using cosine similarity. A siamese architecture with a bi-encoder reduces the computational overhead while improving the quality of representations, compared to a cross-encoder to determine sentence similarity, as in BERT \cite{reimers2019sentencebert}. 

Additionally, SBERT provided significant improvements for semantic text similarity. Although the authors fine-tuned the SBERT model on natural language inference (NLI) data and also applied the model to semantic textual similarity (STS) task, SBERT demonstrated competitive results when being used as a text vectorization method for classification tasks \cite{thuma2023benchmarking}.

\subsubsection{Fine-tuning, data preparation, and loss functions}
\label{subsubsec:fine-tuning}
SBERT allows several strategies for the fine-tuning process. Typically, SBERT requires texts and a label. Due to the siamese characteristic of SBERT, generally, it requires text pairs (or triplets) and a label. Depending on the strategy, this label can correspond to a class, relatedness degree between texts, or relatedness class between texts, e.g., contradiction, neural, or entailment.
The loss functions and respective strategies allowed by SBERT include: 
\begin{itemize}
    \item  \textbf{Batch All Triplets loss (BATL)} \cite{hermans2017defense}, which requires single texts and their respective classes. Internally, same-class texts are treated as positive anchors, while distinct-class texts are considered negative anchors;
    \item \textbf{Cosine Similarity loss (CSL)}, which expects text pairs and a cosine similarity score as a label. A clear drawback is the demand for a cosine similarity, which requires an extra method/ground truth;
    \item \textbf{Contrastive Tension loss (CTL)} \cite{carlsson2020semantic}, which receives single texts without labels. In this case, exact texts are treated as positive anchors, and all the texts are randomly mixed to generate negative anchors. It uses a ratio to define the number of negative anchors for each positive anchor;
    \item \textbf{Multiple Negative Ranking loss (MNRL)} \cite{henderson2017efficient} uses only positive text pairs (or triplets with a negative anchor appended) without label. In this case, texts are mixed to generate negative anchors. A noticeable characteristic is the need for a ground truth to indicate positive texts;
    \item \textbf{Online Contrastive loss (OCL)} requires text pairs and a label indicating their relationship. In this case, the loss function is calculated per item; 
    \item \textbf{Softmax loss (SL)} receives text pairs and a label indicating their relationship. Reimers and Gurevych \cite{reimers2019sentencebert} leveraged this loss function for natural language inference, and therefore, the possible labels were contradiction, neutral, or entailment.
\end{itemize}

The list above is non-exhaustive. This paper aims to evaluate text-based sampling methods (see Section \ref{sec:text-based-sampling-methods}) to gather more useful texts for fine-tuning. 
Considering the above loss functions, we selected BATL, CTL, OCL, and SL. We did not select CSL and MNRL because CSL depends on a similarity measure for the text pairs and would require extra information for this calculation. MNRL, on the other hand, requires positive pairs, or triplets, with a negative anchor. Although we could leverage the classes to generate positive pairs, choosing good-quality anchors can be challenging and require deeper analysis for an assertive selection.


The high cost of training language models is well-known \cite{sharir2020cost}. Although fine-tuning is cheaper than training from scratch, using all new data can also lead to high costs \cite{amba2021dynamic,sharir2020cost}. Thus, resorting to sampling methods can be beneficial in two aspects: (a) selecting more informative texts and (b) consuming fewer computational resources than using all new data.

\section{Text-based Sampling Methods}
\label{sec:text-based-sampling-methods}
This section presents the sampling methods used in this paper for selecting texts for fine-tuning purposes. We also propose the WordpieceToken ratio and later compare it to other text-based sampling methods.

In addition to each sampling method, except for the random sampling, we evaluate an extra scenario leveraging the text labels (classes). Therefore, the sampling methods correspond to their original version and the version that accounts for the class. Algorithm \ref{alg:estimated-error-sampling} provides the weighted sampling pseudocode. We highlight that, optionally, the observed classes' frequencies can be used as an argument for the \texttt{WeightedSampling} function. These frequencies can also be calculated directly from the buffer, using the attribute \textit{class}. The algorithm runs according to the following steps: (1) the buffer containing the stored items from the stream is iterated; (2) each item has its weight calculated, depending on the chosen sampling method; (3) if the classes' frequencies are considered, then the items' probabilities of less frequent classes are increased proportionally; (4) the weights are normalized; and (5) $n_s$ instances are sampled from the buffer.

\begin{algorithm}
\caption{Algorithm of weighted sampling}
\label{alg:estimated-error-sampling}
\begin{algorithmic}
\Require $n_s$ \Comment{number of instances to be sampled}
\Require $\texttt{classes\_frequencies}$ \Comment{observed classes frequencies}
\Require $\texttt{buffer}$ 
\Ensure $\texttt{buffer} \neq \emptyset$

\Function{WeightedSampling}{$n_s, \texttt{buffer}, \texttt{classes\_frequencies}$: Optional}

\ForEach{$X_i$ $\in$ \texttt{buffer}}
    \State{\textrm{Calculate $X_i.\texttt{weight}$ according to the sampling method}}
    \If{$\texttt{classes\_frequencies} \neq \texttt{None}$}
        \State $X_i.\texttt{weight}^{*} \gets X_i.\texttt{weight} \times \frac{\texttt{sum}(\texttt{classes\_frequencies})}{\texttt{classes\_frequencies}[X_i.\texttt{class}]}$
    \EndIf
\EndFor
\State{\textrm{Normalize weights}}

\State{\textrm{Sample $n_s$ instances using the calculated $X_i.\texttt{weight}^{*}$ as probability}}
\EndFunction
\end{algorithmic}
\end{algorithm}

The text sampling methods employed in this paper are:
\begin{itemize}
    \item \textbf{Length-based sampling} \cite{amba2021dynamic}: it ponders items by their length. This means we counted the number of tokens in each text and normalized them using the biggest and the lowest lengths. The main idea is that longer texts have more chance to encompass useful, novel information for the language model in the fine-tuning process;

    \item \textbf{Random sampling}: this sampling method randomly selects a given number of items from the buffer;

    \item \textbf{TF-IDF-based sampling}: Term Frequency - Inverse of Document Frequency (TF-IDF) \cite{salton1988term} is a technique for measuring the importance of a given word in a document, considering a collection. The term frequency is the count of a token $t$ in a text $d$. The inverse of document frequency is generally calculated as $\textrm{idf}(t) = \log{\frac{n}{\textrm{document\_frequency}(t)}}$, where $\textrm{document\_frequency}(t)$ corresponds to the number of documents containing $t$, and $n$ is the total number of documents. Then, the complete TF-IDF calculation is: $\textrm{TFIDF}(t,d) = \textrm{term\_frequency}(t,d) \times \textrm{idf}(t)$. Given a text, the TF-IDF is calculated for each token in the text. Thus, the text's weight is the sum of the TF-IDF values of the tokens present in the text. The rationale behind this approach is to select texts with more important words across the buffer.
\end{itemize}

    \review{In addition to these sampling methods, we propose a novel text sampling method named WordpieceToken ratio.}


\vspace{-15pt}    
\review{\subsection{WordpieceToken ratio sampling}}

    \review{This paper proposes a novel sampling method based on the ratio between wordpieces and tokens. Wordpiece is a technique BERT-based models use to handle out-of-vocabulary (OOV) tokens \cite{devlin2019bert}. 
    For example, the word \texttt{institutionalization} could be partitioned into two wordpieces: \texttt{[`institutional',  `\#\#ization']}, where \texttt{\#\#} means that there is a previous partition.}
    
    \review{The rationale behind this sampling method is that the bigger the ratio between wordpieces and tokens, the bigger the number of unknown words (by the language model) in the text. The ratio is calculated per instance. For example, let the text instance be: ``Extramedullary toxicity was limited to hypothyroidism'', a 6-token sentence. When applied to a BERT tokenizer, it is split into \texttt{[`extra', `\#\#med', `\#\#ulla', `\#\#ry', `toxicity', `was', `limited', `to', `h', `\#\#yp', `\#\#oth', `\#\#yr', `\#\#oid', `\#\#ism']}, resulting in 14 wordpieces. Thus, this instance's weight is $\frac{14}{6} = 2.33$. After all the text instances from the buffer have their weights calculated, several instances are sampled considering the instances' weights and class frequencies according to Algorithm \ref{alg:estimated-error-sampling}.}
    
    \review{Therefore, we hypothesize that sampling texts by weighting the WordpieceToken ratio may retrieve texts with more useful information for the fine-tuning process.}

\section{Experimental Results}
\label{sec:experimental-results}

This section provides the experimental results, including the experimental protocol, datasets, proposed scenario, evaluation metrics, and the results.

\subsection{Experimental Protocol}

\subsubsection{Datasets}
Two datasets were used: Airbnb and Yelp. 

\begin{itemize}
    \item \textbf{Airbnb:} This dataset was obtained from the Inside Airbnb\footnote{http://insideairbnb.com/get-the-data}, considering data related to the New York City. Since New York City is one of the most popular destinations in the United States\footnote{Available at: https://www.cntraveler.com/story/most-visited-american-cities. Accessed on Jan 20th, 2024.}, the Airbnb dataset related to New York City has several reviews, enough for multiple sampling for text stream simulation (see Section \ref{subsec:proposed-scenario}). This dataset, by default, is not ready for classification tasks. Therefore, we leveraged: (a) a pre-trained model for language identification is used (lid.176.ftz \cite{joulin2016bag,joulin2016fasttext}), and (b) a pre-trained model for sentiment analysis to infer the reviews' sentiment (Twitter RoBERTa Base Sentiment\footnote{Available at: https://huggingface.co/cardiffnlp/twitter-roberta-base-sentiment.} model \cite{barbieri2020tweeteval}). Thus, the English reviews were filtered, and their sentiments were inferred. The sentiments, i.e., positive, negative, and neutral, were used as labels in the classification task.
    The processing steps are available on Github\footnote{https://github.com/cristianomg10/methods-for-generating-drift-in-text-streams}.
    \item \textbf{Yelp:} This dataset is provided on Yelp Datasets\footnote{https://www.yelp.com/dataset}. Yelp consists of reviews collected regarding over 130 thousand businesses. These reviews are accompanied by a category from a scale of stars between 1 and 5. This category is used as a label in the text classification task.
\end{itemize}

\begin{figure*}[htp!]
    \centering
    \begin{subfigure}[t]{0.43\textwidth}
        \centering
        \includegraphics[width=\linewidth]{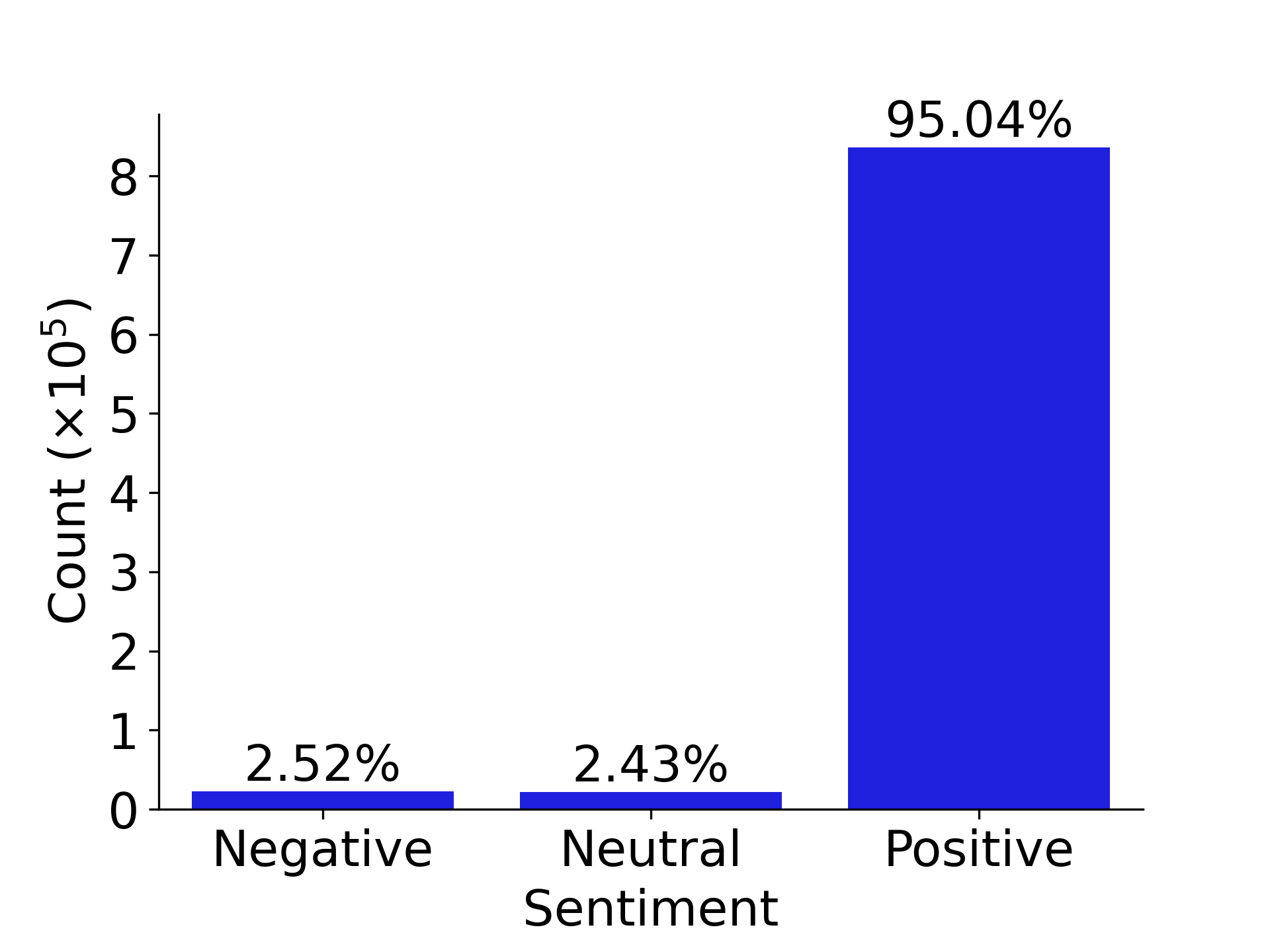}
        \caption{Class distribution of Airbnb dataset after filtering.}
        \label{fig:airbnb-dataset-distr}
    \end{subfigure}%
    ~ 
    \begin{subfigure}[t]{0.43\textwidth}
        \centering
        \includegraphics[width=\linewidth]{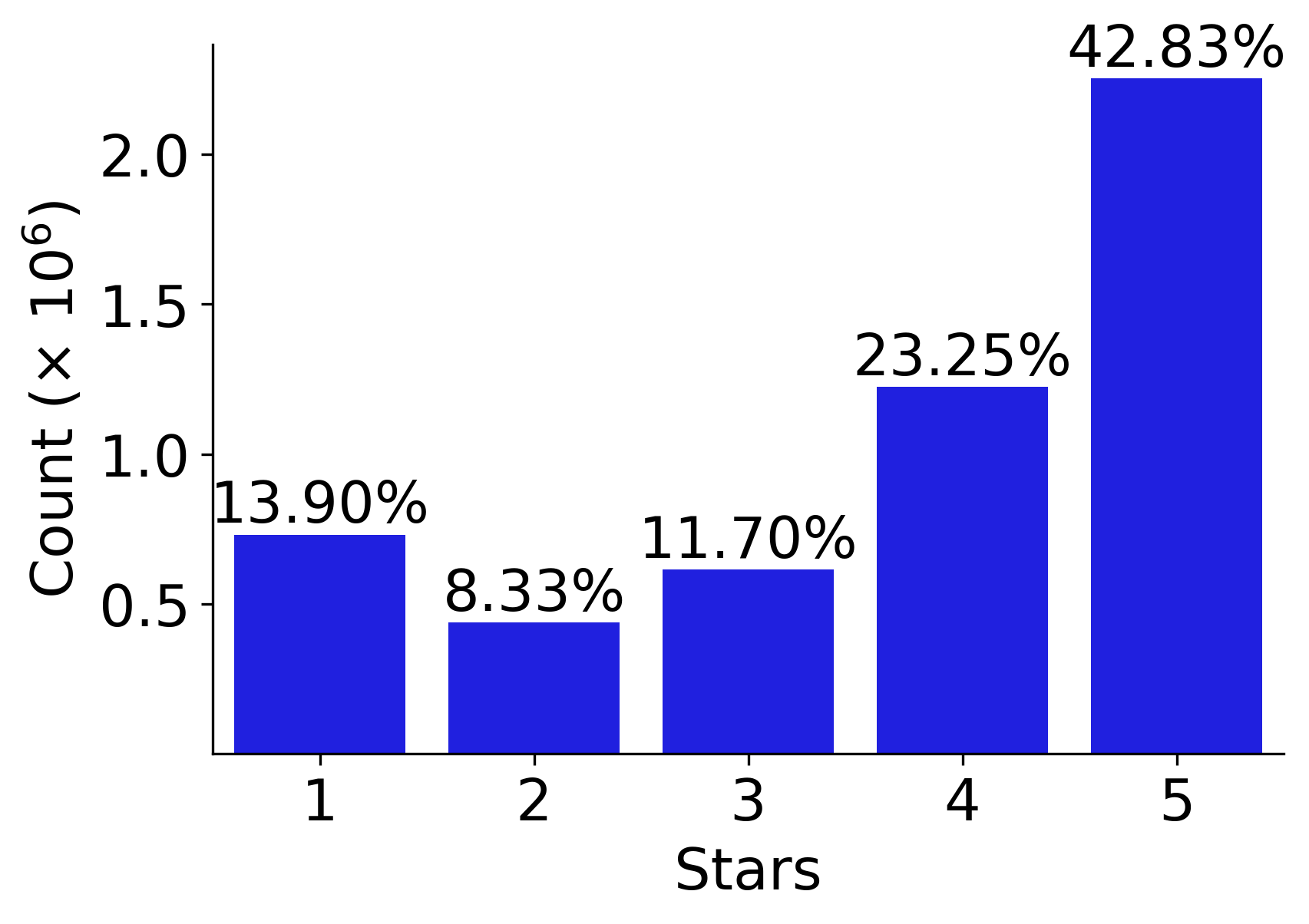}
        \caption{Class distribution of Yelp dataset. Adapted from \cite{thuma2023benchmarking}.}
        \label{fig:yelp-dataset-distr}
    \end{subfigure}%
    \caption{Class distribution of the datasets used in this paper.}
    \label{fig:distributions}
\end{figure*}

An essential characteristic regarding the above datasets is the presence of a timestamp, which is crucial for text stream simulation. The data distributions are presented in Fig. \ref{fig:distributions}. Noticeably, the datasets are imbalanced, reflecting the nature of real-world data.

Other relevant information on the experiments are:
\begin{itemize}
    \label{sss:sample_sizes}
    \item \textbf{Sample sizes:} we considered four sample sizes: 500, 1000, 2500, and 5000. Those values were chosen since they represent between 1\% and 10\% of the buffer size, which can be considered reasonable values;
    \item \textbf{Classifier:} \review{we selected the Incremental Support Vector Machine (ISVM) as the classifier because it is updated incrementally according to the arrival of new data. As its batch counterpart \cite{vapnik1995support}, ISVM calculates an optimal hyperplane between instances of two distinct classes, and the surrounding instances are assumed as support vectors \cite{d2019monitoring}.}
    \item \textbf{Hardware:} the hardware used in the experiments is a 13th Gen Intel(R) Core(TM) i9-13900K, 128 GB of RAM running Ubuntu 22.04 LTS, and 2 x GPU GeForce RTX 4090 (24GB). 
\end{itemize}





\subsection{Proposed Scenario}
\label{subsec:proposed-scenario}
Considering the text stream mining setting, the proposed scenario is run as follows: (1) a text stream of length 200,000 sampled (stratified by class/label) from the original datasets; (2) the text stream classification is performed one-by-one; (3) a buffer accumulates the first 50,000 items of the text stream; (4) at the moment $t=50,000$, a sampling method, among the methods presented in this paper, will sample a predefined number of items from those described in Section \ref{sss:sample_sizes}; (5) after sampling, the fine-tuning process for the language model, i.e., the SBERT model, is triggered using the sampled texts and the selected loss function; and (6) the evaluation metrics are calculated cumulatively in a test-then-train fashion, i.e., in a prequential manner.


This process was executed five times per sampling method and loss function.
In each run, the entire stream was also sampled from the original dataset in a stratified manner. 
We used $t=50,000$ as the effects of the fine-tuning and sampling methods would be easier to spot than at the end of the stream.

\subsection{Loss Functions Settings}
Considering the loss functions presented in Section \ref{subsubsec:fine-tuning}, their inputs for fine-tuning were defined as follows:


\begin{itemize}
    \item \textbf{BATL:} single texts and their respective classes;
    \item \textbf{CTL:} single texts;
    \item \textbf{OCL:} text pairs and a label, which we adapted for considering the distance between classes by calculating: $\textrm{label} = 1 - \frac{abs(X_1.\textrm{class} - X_2.\textrm{class})}{|classes|}$, where $X_{\phi}$ is a text, $|\cdot|$ is the cardinality, and $\textrm{abs}(\cdot)$ is the absolute value. Thus, $\textrm{label}$ is one if the classes are the same, indicating the similarity;
    \item \textbf{SL}: similarly to OCL, SL receives text pairs and a label, which we adapted to be the absolute distance, calculated as $\textrm{label} = \textrm{abs}(X_1.\textrm{class} - X_2.\textrm{class})$, where $X_\phi$ is a text, and $\textrm{abs}(\cdot)$ is the absolute value.
\end{itemize}

This paper leverages the pre-trained SBERT model \texttt{paraphrase-MiniLM-L6-v2}. When fine-tuning, we used 32-sized batches, 10 epochs, and 100 warmup steps, which are values frequent in the documentation.


\subsection{Evaluation Metrics}

Given the class imbalance of the datasets used in experimentation, results regarding Macro F1 Score and elapsed time were reported. In particular, the Macro F1 Score averages the harmonic mean of precision and recall obtained per class. 


\subsection{Results}
\begin{landscape}

\begin{figure}[!htp]
\centering
\includegraphics[width=\linewidth]{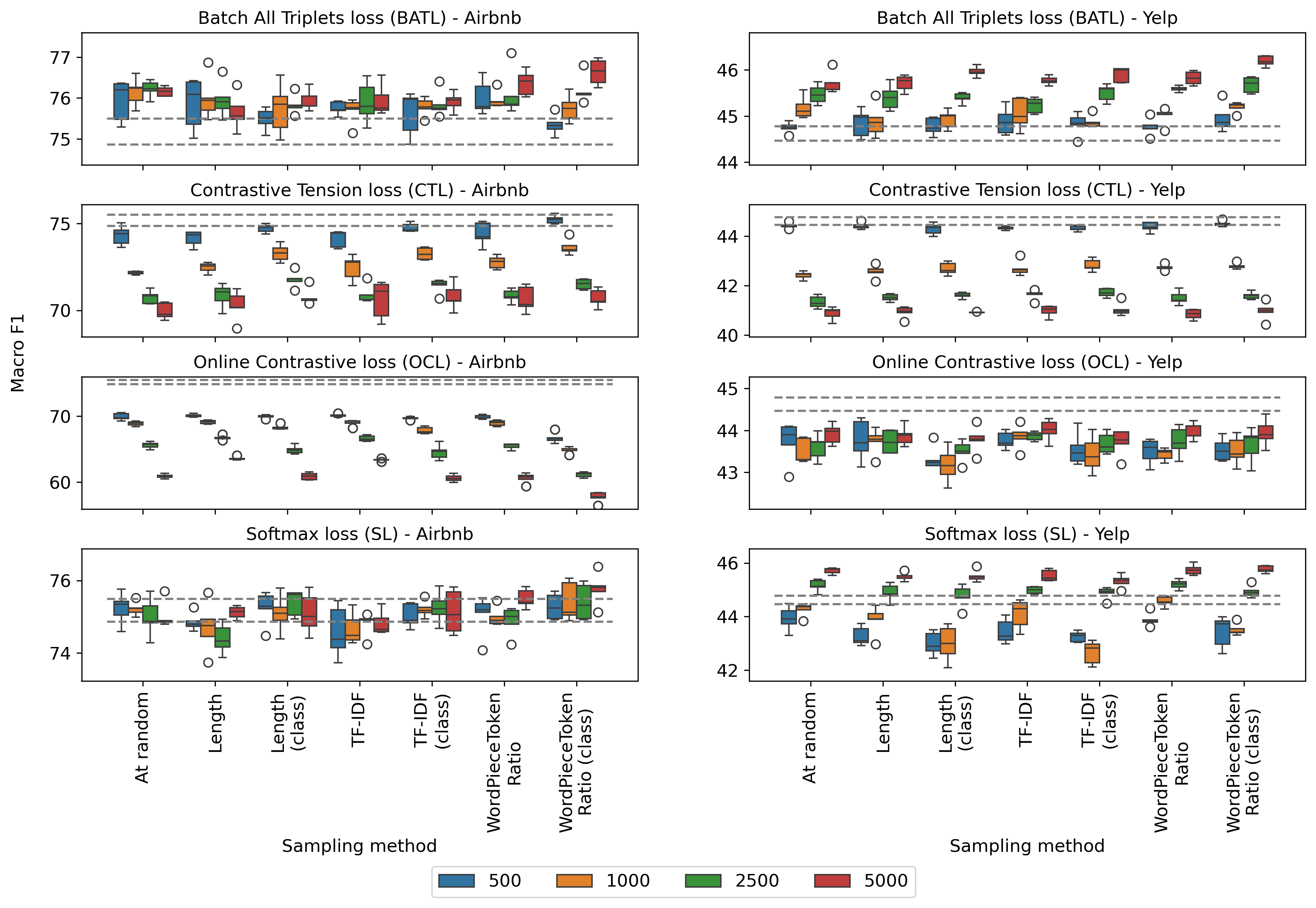}
\caption{Results for Airbnb (left) and Yelp (right). The dashed lines correspond to the maximum and minimum Macro F1-Score values using SBERT without update. The numbers in the legend correspond to the sample sizes.}
\label{fig:results}
\end{figure}

\end{landscape}

Considering the presented scenario, loss functions, and sampling methods, the results obtained regarding Macro F1-Scores are demonstrated in Fig. \ref{fig:results}. For readability issues, we kept different scales on the y-axis. The dashed lines correspond to the maximum and minimum Macro F1-Score values using SBERT without update and were used as a baseline. The x-axis regards the sampling methods, while the boxes correspond to the tested sample sizes. 
The Macro F1-Scores obtained for the baseline were: (a) 75.13 $\pm$ 0.28 (min: 74.86; max: 75.50) for the Airbnb dataset, and (b) 44.60 $\pm$ 0.12 (min: 44.46; max: 44.78) for Yelp dataset. Regarding the elapsed times, the values obtained were (in seconds): (a) 496.78 $\pm$ 0.70 (min: 495.89; max: 497.86) for the Airbnb dataset, and (b) 558.82 $\pm$ 3.19 (min: 553.36; max: 561.10) for Yelp dataset.

Considering the results obtained for the Airbnb dataset, we can see that the loss function is crucial for improving the results. CTL and OCL performed worse than when there were no updates, according to the dashed lines. Considering OCL and CTL, only the proposed WordPieceToken ratio sampling method (using class in the sampling, sample size 500, and CTL) performed equivalently to SBERT without update. Apart from this point, all the combinations for CTL and OCL performed worse than SBERT without update. Although sometimes equivalent to the competitors, the proposed WordPieceToken ratio (class) generally obtains higher averages than its peers with the same sample size, except for using the OCL. Furthermore, increasing the sample size for these loss functions degraded the results. We assume that these loss functions, together with the looseness of anchoring, ease the model to suffer from catastrophic forgetting.

Still analyzing the results for the Airbnb dataset, regarding BATL and SL, mostly in SL, all sampling methods were equivalent to SBERT without updates. However, using the Length or TF-IDF sampling method without accounting for the classes led to smaller Macro F1-Scores. In addition, it is possible to notice that, in the SL case, accounting for the classes in the sampling method helped reach a better Macro F1-Score than the same sampling method without accounting for the class. In the BATL scenario, all methods perform similarly across the sample sizes, except for the WordPieceToken ratio, which obtained the highest Macro F1-Score using a sample size of 5000. Therefore, using a sample size of 5000 improved the performance in these cases.

Similar observations can be made by switching to the results concerning the Yelp dataset: CTL and OCL led to poorer results than BATL and SL. CTL provided the worst results for this dataset; the bigger the sample size, the worse the performance. Again, CTL seems to lead to catastrophic forgetting. It can be credited to its simple way of generating text pairs for fine-tuning. On the other hand, OCL obtained equivalent results across sampling methods and sample sizes, but all were worse than SBERT without updates. In addition, considering the sample sizes of 2500 and 5000, all methods showed increased performance.

BATL and SL functions led to increased performance for the Yelp dataset compared to the baseline (dashed lines). For BATL, all sampling methods using the sample size of 500 and 1000 are equivalent to the baseline. From the sample size of 1000, the At Random and WordPieceToken ratio sample methods reached Macro F1-Score values superior to the baseline. Regarding SL, most sampling methods are superior to the baseline from the sample size of 2500. Smaller samples led to decreased performance compared to the baseline.

\begin{landscape}

\begin{figure}[!htp]
\centering
\includegraphics[width=\linewidth]{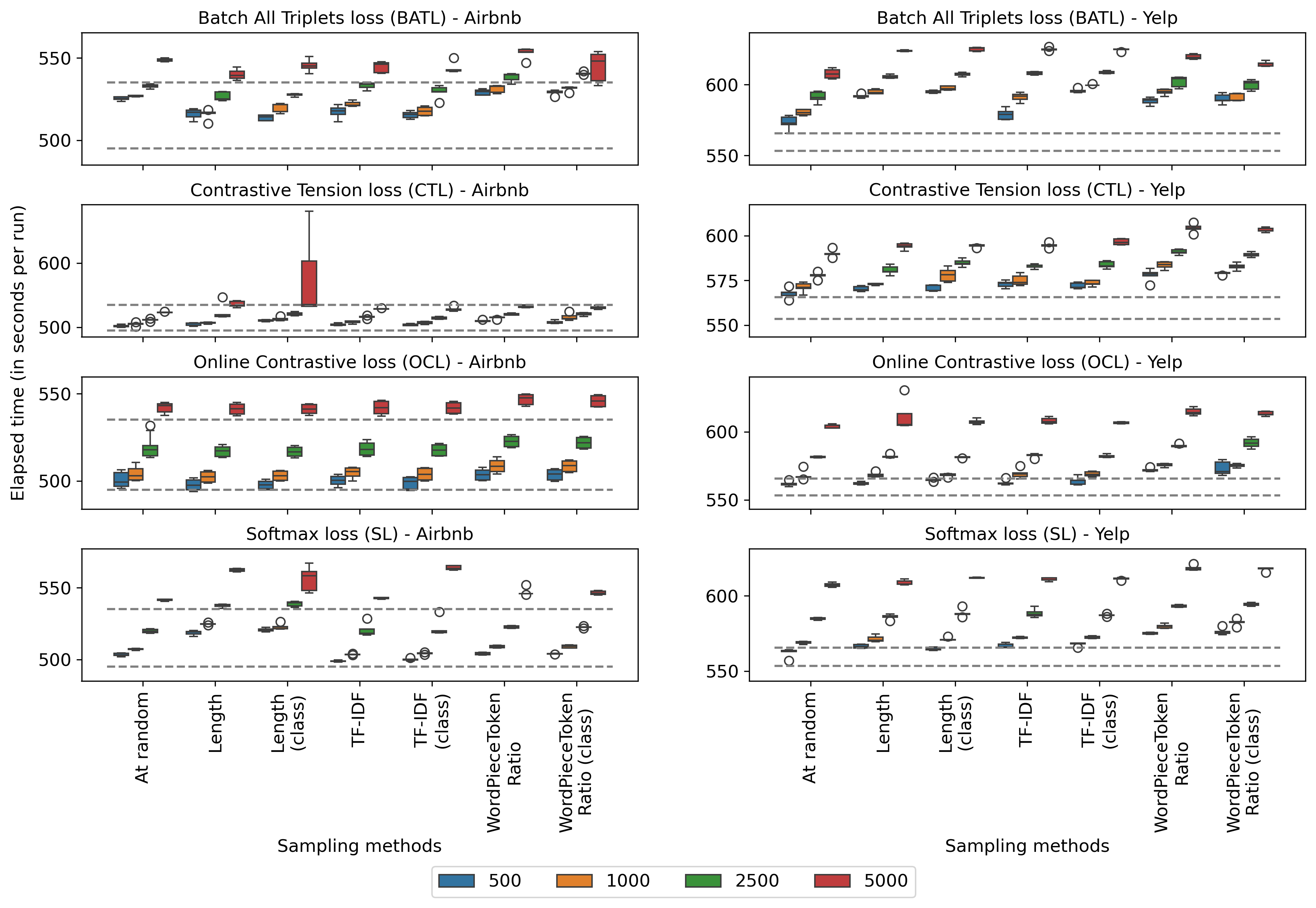}
\caption{Elapsed times for Airbnb (left) and Yelp (right). The dashed lines correspond to the maximum and minimum Macro F1-Score values using SBERT without update.}
\label{fig:elapsed-times}
\end{figure}

\end{landscape}

Regarding the elapsed times, Fig. \ref{fig:elapsed-times} shows measured values in each setting. Again, dashed lines correspond to the baseline, i.e., SBERT without update. For the Airbnb dataset using the BATL function, we noticed that the proposed WordPieceToken ratio took longer than the baseline to run, essentially from the sample size of 2500. For CTL, all methods, except for the length sampling method in both variations, took longer than the baseline. Specifically for Length (class), it was unstable, taking a reasonable time when using the sample size of 5000. It somehow can be expected since longer reviews would be selected, and the fine-tuning process would take longer. However, this should also be visualized for the Length sampling method, but it has not happened. We hypothesize that an anomaly in the GPU may have led to this increased variation. In OCL, all sampling methods have similar elapsed times: only sampling 5000 items led to higher run times than the baseline. At last, for SL, the Length- and TF-IDF-based methods took longer than the baseline from the sample size of 2500. In addition, Length (class) showed similar behavior to CTL regarding instability.

Regarding the Yelp dataset, considering the BATL function, random sampling led to shorter elapsed times. Length- and TF-IDF-based (with class) sampling methods led to longer elapsed times. Differently, in CTL, the proposed WordPieceToken ratio sampling reached the highest run times compared to all other methods. The same behavior happens in the SL function. For the OCL, the elapsed times for all methods are equivalent to the ones from SBERT without update, except for the WordPieceToken ratio sampling method.


Table \ref{tab:results} condenses the results obtained. The bold values are the best per dataset/sample size/loss function (LF) combination, i.e., per row. The values in yellow and green are the best Macro F1 scores and elapsed times per dataset/sample size. WordPieceToken ratio (class) obtained the best Macro F1-Scores in 5 out of 8 dataset/sample size pairs, and WordPieceToken ratio in 1 out of 8. Sampling at random was the fastest in 6 out of 8 dataset/sample size pairs. Furthermore, the best Macro F1-Scores increase with the sample size.

Although elapsed time is important and variations followed similar patterns, i.e., the higher the sample size, the longer the elapsed time, the differences between them show that they may not impede fine-tuning in text stream scenarios, depending on the hardware. 
The differences between maximum and minimum elapsed times are 187.07 seconds for Airbnb and 73.54 seconds for Yelp, which represent 37\% and 13\% of their respective average elapsed times. 

\section{Conclusion}
\label{sec:conclusion}

Learning from text streams is challenging due to the constraints of text streams, such as time, resources, and one-pass processing \cite{gama2014survey,garcia2023concept}. Furthermore, the existence of concept drifts in texts produced over time is well-known. A way to overcome textual concept drifts is by updating the language model. Updating (or fine-tuning) the language model is generally costly if all new data is considered.

This paper evaluates four different sampling methods, i.e., random sampling, length sampling, TF-IDF sampling, and WordPieceToken ratio sampling (proposed in this paper), where the latter three had a version that accounted for instances' classes, aiming at sampling important, informative texts from the buffer.  Four loss functions were assessed in combination with the text sampling methods, i.e., Batch All Triplets loss (BATL), Contrastive Tension loss (CTL), Online Contrastive loss (OCL), and Softmax loss (SL).

\review{We observed that the loss function plays a crucial role in improving the performance of the text classification task. Considering the scenarios assessed, CTL and OCL functions were insufficient to achieve satisfactory performance levels.} On the other hand, BATL and SL functions aided in maintaining interesting performance levels, sometimes above the SBERT without update (our baseline), suggesting the SBERT can benefit from these loss functions. In addition, the elapsed times were comparable to the baseline. Therefore, BATL and SL were the most suitable functions for text stream classification among those evaluated in this paper. However, one must be careful when employing this approach because fine-tuning can create a bottleneck in real-time applications. Finally, text sampling methods are also crucial in fine-tuning. Our experiments suggest that the proposed WordPieceToken ratio method, especially leveraging the instances' classes, can retrieve more informative texts and favor the machine learning model's performance after fine-tuning. 

\begin{landscape}

\begin{table}[]
\caption{Condensed results. Bold values are the best values obtained per row. Values in yellow (Macro F1) and green (elapsed time) are the best per dataset/sample size.}
\label{tab:results}
\resizebox{\linewidth}{!}{
\begin{tabular}{lllllllllllllllll}
\cline{3-16}
                                & \multicolumn{1}{l|}{}                 & \multicolumn{2}{c|}{\textbf{At random}}                                            & \multicolumn{2}{c|}{\textbf{Length}}                                               & \multicolumn{2}{c|}{\textbf{Length (class)}}                                       & \multicolumn{2}{c|}{\textbf{TF-IDF}}                                               & \multicolumn{2}{c|}{\textbf{TF-IDF (class)}}                                       & \multicolumn{2}{c|}{\textbf{WordPieceToken ratio}}                                 & \multicolumn{2}{c}{\textbf{WordPieceT. (class)}}                                  &  \\ \cline{1-16}
\multicolumn{1}{c}{\textbf{Dataset}} & \multicolumn{1}{c|}{\textbf{LF}} & \multicolumn{1}{c}{\textbf{Macro F1}} & \multicolumn{1}{c|}{\textbf{Elaps. time}} & \multicolumn{1}{c}{\textbf{Macro F1}} & \multicolumn{1}{c|}{\textbf{Elaps. time}} & \multicolumn{1}{c}{\textbf{Macro F1}} & \multicolumn{1}{c|}{\textbf{Elaps. time}} & \multicolumn{1}{c}{\textbf{Macro F1}} & \multicolumn{1}{c|}{\textbf{Elaps. time}} & \multicolumn{1}{c}{\textbf{Macro F1}} & \multicolumn{1}{c|}{\textbf{Elaps. time}} & \multicolumn{1}{c}{\textbf{Macro F1}} & \multicolumn{1}{c|}{\textbf{Elaps. time}} & \multicolumn{1}{c}{\textbf{Macro F1}} & \multicolumn{1}{c}{\textbf{Elaps. time}} &  \\ \cline{1-16}
Airbnb & BATL & 75.94$\pm$0.51 & 525.29$\pm$1.09 & 75.86$\pm$0.63 & 516.03$\pm$3.20 & 75.49$\pm$0.27 & \textbf{513.74$\pm$1.74} & 75.75$\pm$0.16 & 517.22$\pm$3.97 & 75.63$\pm$0.55 & 515.42$\pm$2.17 & \cellcolor{yellow!75}\textbf{76.01$\pm$0.43}  & 529.26$\pm$1.71 & 75.34$\pm$0.25 & 529.00$\pm$1.55 \\
 (500) & CTL & 74.31$\pm$0.57 & \textbf{502.06$\pm$1.84} & 74.13$\pm$0.45 & 504.31$\pm$2.31 & 74.72$\pm$0.24 & 510.63$\pm$1.34 & 74.12$\pm$0.48 & 504.55$\pm$1.81 & 74.76$\pm$0.25 & 503.81$\pm$1.52 & 74.39$\pm$0.68 & 510.04$\pm$1.32 & \textbf{75.24$\pm$0.24} & 508.18$\pm$2.29 \\
 & OCL & 69.92$\pm$0.49 & 500.51$\pm$4.21 & \textbf{70.13$\pm$0.22} & 497.78$\pm$3.27 & 69.93$\pm$0.26 & \cellcolor{green!75}\textbf{497.78$\pm$2.52} & 70.12$\pm$0.21 & 500.30$\pm$2.81 & 69.70$\pm$0.21 & 498.81$\pm$3.43 & 69.95$\pm$0.28 & 503.60$\pm$3.05 & 66.71$\pm$0.74 & 503.60$\pm$3.23 \\
 & SL & 75.24$\pm$0.44 & 503.34$\pm$1.12 & 74.86$\pm$0.25 & 518.56$\pm$1.62 & 75.25$\pm$0.47 & 520.60$\pm$1.19 & 74.58$\pm$0.72 & \textbf{498.92$\pm$0.48} & 75.03$\pm$0.33 & 499.92$\pm$0.75 & 75.06$\pm$0.57 & 504.14$\pm$0.86 & \textbf{75.29$\pm$0.36} & 503.97$\pm$0.27 \\\hline
Airbnb & BATL & \cellcolor{yellow!75}\textbf{76.15$\pm$0.35} & 526.95$\pm$0.48 & 76.00$\pm$0.53 & \textbf{515.73$\pm$3.28} & 75.74$\pm$0.62 & 519.87$\pm$2.88 & 75.70$\pm$0.32 & 522.05$\pm$1.64 & 75.79$\pm$0.22 & 517.72$\pm$2.71 & 75.95$\pm$0.21 & 531.35$\pm$2.33 & 75.75$\pm$0.33 & 531.44$\pm$1.53 \\
(1000) & CTL & 72.16$\pm$0.09 & \textbf{505.08$\pm$2.57} & 72.46$\pm$0.29 & 507.00$\pm$1.24 & 73.30$\pm$0.49 & 512.99$\pm$2.82 & 72.44$\pm$0.73 & 508.19$\pm$2.09 & 73.25$\pm$0.35 & 507.29$\pm$1.95 & 72.76$\pm$0.37 & 515.10$\pm$1.81 & \textbf{73.64$\pm$0.45} & 516.06$\pm$5.47 \\
 & OCL & 68.91$\pm$0.32 & 503.90$\pm$3.84 & \textbf{69.12$\pm$0.25} & \cellcolor{green!75}\textbf{502.31$\pm$3.06} & 68.31$\pm$0.37 & 502.92$\pm$2.82 & 68.97$\pm$0.43 & 504.76$\pm$3.00 & 67.86$\pm$0.49 & 503.85$\pm$3.45 & 68.96$\pm$0.39 & 508.66$\pm$3.74 & 64.88$\pm$0.46 & 508.48$\pm$3.26 \\
 & SL & 75.23$\pm$0.20 & 507.20$\pm$0.53 & 74.71$\pm$0.70 & 524.81$\pm$0.69 & 75.09$\pm$0.51 & 522.96$\pm$2.10 & 74.67$\pm$0.44 & \textbf{503.55$\pm$0.41} & 75.21$\pm$0.23 & 504.27$\pm$0.56 & 75.00$\pm$0.27 & 508.90$\pm$0.87 & \textbf{75.42$\pm$0.55} & 509.07$\pm$1.04 \\\hline
Airbnb  & BATL & \cellcolor{yellow!75}\textbf{76.23$\pm$0.21} & 532.92$\pm$1.24 & 75.96$\pm$0.44 & \textbf{526.52$\pm$2.68} & 75.84$\pm$0.24 & 527.48$\pm$0.85 & 75.90$\pm$0.51 & 533.20$\pm$2.16 & 75.85$\pm$0.33 & 529.35$\pm$4.03 & 76.10$\pm$0.57 & 538.23$\pm$2.76 & 76.20$\pm$0.35 & 540.58$\pm$0.78 \\
(2500) & CTL & 70.76$\pm$0.39 & \cellcolor{green!75}\textbf{511.41$\pm$1.76} & 70.84$\pm$0.68 & 523.80$\pm$12.93 & \textbf{71.78$\pm$0.47} & 520.68$\pm$2.58 & 70.96$\pm$0.52 & 516.08$\pm$1.99 & 71.41$\pm$0.43 & 514.46$\pm$1.72 & 70.84$\pm$0.38 & 520.30$\pm$1.42 & 71.51$\pm$0.29 & 520.60$\pm$2.34 \\
 & OCL & 65.61$\pm$0.47 & 519.29$\pm$6.45 & \textbf{66.71$\pm$0.31} & 516.95$\pm$2.99 & 64.91$\pm$0.59 & \textbf{516.74$\pm$2.84} & 66.63$\pm$0.41 & 518.33$\pm$3.89 & 64.56$\pm$1.05 & 517.66$\pm$3.32 & 65.46$\pm$0.42 & 522.70$\pm$3.29 & 61.16$\pm$0.37 & 521.83$\pm$3.31 \\
 & SL & 75.00$\pm$0.53 & \textbf{519.83$\pm$1.29} & 74.40$\pm$0.42 & 537.47$\pm$1.17 & 75.39$\pm$0.36 & 538.80$\pm$1.83 & 74.82$\pm$0.33 & 520.59$\pm$4.74 & 75.25$\pm$0.44 & 521.87$\pm$6.34 & 74.89$\pm$0.40 & 522.76$\pm$0.90 & \textbf{75.41$\pm$0.50} & 522.59$\pm$0.59 \\\hline
Airbnb  & BATL & 76.16$\pm$0.12 & 548.77$\pm$0.93 & 75.66$\pm$0.44 & \textbf{539.97$\pm$3.28} & 75.94$\pm$0.26 & 545.38$\pm$3.80 & 75.95$\pm$0.38 & 544.53$\pm$3.41 & 75.91$\pm$0.23 & 543.80$\pm$3.49 & 76.37$\pm$0.31 & 552.95$\pm$3.40 & \cellcolor{yellow!75}\textbf{76.64$\pm$0.32} & 544.69$\pm$9.47 \\
(5000) & CTL & 69.94$\pm$0.47 & \cellcolor{green!75}\textbf{523.54$\pm$0.68} & 70.27$\pm$0.86 & 536.06$\pm$4.73 & 70.77$\pm$0.49 & 577.01$\pm$65.48 & 70.62$\pm$1.10 & 528.92$\pm$0.60 & \textbf{70.81$\pm$0.78} & 528.31$\pm$3.22 & 70.63$\pm$0.74 & 532.11$\pm$1.80 & 70.70$\pm$0.53 & 530.71$\pm$2.38 \\
 & OCL & 60.90$\pm$0.29 & 541.87$\pm$2.84 & 63.61$\pm$0.24 & 541.25$\pm$3.14 & 60.90$\pm$0.52 & \textbf{541.18$\pm$2.90} & \textbf{63.37$\pm$0.18} & 541.97$\pm$3.90 & 60.62$\pm$0.50 & 541.88$\pm$3.27 & 60.61$\pm$0.73 & 546.76$\pm$2.96 & 57.73$\pm$0.75 & 545.78$\pm$3.48 \\
 & SL & 75.02$\pm$0.38 & \textbf{541.39$\pm$0.54} & 75.12$\pm$0.17 & 562.32$\pm$1.08 & 75.10$\pm$0.57 & 556.26$\pm$8.92 & 74.83$\pm$0.34 & 542.58$\pm$0.51 & 75.13$\pm$0.61 & 563.87$\pm$1.56 & 75.50$\pm$0.27 & 546.98$\pm$2.86 & \textbf{75.77$\pm$0.45} & 546.37$\pm$1.39 \\\hline
Yelp  & BATL & 44.75$\pm$0.12 & \textbf{573.14$\pm$5.07} & 44.85$\pm$0.31 & 591.92$\pm$1.27 & 44.78$\pm$0.19 & 594.89$\pm$0.87 & 44.89$\pm$0.30 & 579.11$\pm$3.89 & 44.83$\pm$0.25 & 595.63$\pm$1.35 & 44.77$\pm$0.19 & 588.47$\pm$2.56 & \cellcolor{yellow!75}\textbf{44.96$\pm$0.31} & 590.06$\pm$3.50 \\
(500) & CTL & 44.43$\pm$0.11 & \textbf{567.70$\pm$2.88} & 44.42$\pm$0.14 & 570.24$\pm$1.42 & 44.30$\pm$0.23 & 571.04$\pm$1.79 & 44.33$\pm$0.07 & 572.89$\pm$1.88 & 44.32$\pm$0.11 & 571.94$\pm$1.63 & 44.37$\pm$0.19 & 577.88$\pm$3.47 & \textbf{44.51$\pm$0.11} & 578.93$\pm$0.70 \\
 & OCL & 43.73$\pm$0.50 & \cellcolor{green!75}\textbf{561.79$\pm$1.79} & \textbf{43.77$\pm$0.49} & 562.14$\pm$1.05 & 43.33$\pm$0.28 & 564.83$\pm$1.13 & 43.76$\pm$0.21 & 562.65$\pm$2.03 & 43.55$\pm$0.39 & 564.05$\pm$2.98 & 43.50$\pm$0.31 & 572.15$\pm$1.28 & 43.54$\pm$0.27 & 573.24$\pm$5.18 \\
 & SL & \textbf{43.93$\pm$0.46} & \textbf{562.21$\pm$2.95} & 43.27$\pm$0.35 & 566.43$\pm$1.35 & 42.98$\pm$0.44 & 564.91$\pm$1.02 & 43.45$\pm$0.46 & 567.24$\pm$1.46 & 43.25$\pm$0.20 & 567.83$\pm$1.32 & 43.89$\pm$0.25 & 575.07$\pm$0.55 & 43.43$\pm$0.60 & 576.29$\pm$2.29 \\\hline
Yelp  & BATL & 45.18$\pm$0.25 & \textbf{580.35$\pm$2.03} & 44.89$\pm$0.36 & 594.96$\pm$1.61 & 44.93$\pm$0.20 & 597.51$\pm$1.41 & 45.05$\pm$0.34 & 591.17$\pm$3.14 & 44.87$\pm$0.14 & 599.75$\pm$0.48 & 45.00$\pm$0.18 & 594.88$\pm$2.09 & \cellcolor{yellow!75}\textbf{45.19$\pm$0.11} & 590.76$\pm$2.66 \\
(1000) & CTL & 42.42$\pm$0.16 & \textbf{571.09$\pm$2.78} & 42.56$\pm$0.26 & 572.81$\pm$0.45 & 42.69$\pm$0.25 & 578.06$\pm$3.92 & 42.70$\pm$0.31 & 575.08$\pm$3.13 & \textbf{42.83$\pm$0.25} & 573.45$\pm$1.60 & 42.74$\pm$0.11 & 583.48$\pm$2.05 & 42.79$\pm$0.12 & 582.68$\pm$1.84 \\
 & OCL & 43.50$\pm$0.30 & \cellcolor{green!75}\textbf{568.04$\pm$3.69} & 43.74$\pm$0.31 & 568.48$\pm$1.65 & 43.17$\pm$0.42 & 568.33$\pm$1.18 & \textbf{43.85$\pm$0.29} & 569.32$\pm$3.51 & 43.43$\pm$0.44 & 569.00$\pm$1.84 & 43.43$\pm$0.15 & 575.65$\pm$1.27 & 43.52$\pm$0.34 & 575.20$\pm$1.30 \\
 & SL & 44.27$\pm$0.26 & \textbf{568.90$\pm$0.80} & 43.86$\pm$0.54 & 571.41$\pm$2.11 & 43.00$\pm$0.67 & 571.23$\pm$1.02 & 44.10$\pm$0.55 & 572.35$\pm$0.61 & 42.66$\pm$0.44 & 572.53$\pm$0.84 & \textbf{44.57$\pm$0.20} & 579.77$\pm$1.38 & 43.51$\pm$0.23 & 582.36$\pm$2.09 \\\hline
Yelp  & BATL & 45.47$\pm$0.21 & \textbf{591.20$\pm$3.90} & 45.40$\pm$0.28 & 605.69$\pm$1.19 & 45.39$\pm$0.11 & 607.26$\pm$1.19 & 45.23$\pm$0.17 & 608.03$\pm$1.13 & 45.50$\pm$0.19 & 608.62$\pm$0.88 & 45.59$\pm$0.06 & 602.05$\pm$3.93 & \cellcolor{yellow!75}\textbf{45.68$\pm$0.17} & 599.80$\pm$3.57 \\
(2500) & CTL & 41.34$\pm$0.25 & \cellcolor{green!75}\textbf{577.66$\pm$1.80} & 41.52$\pm$0.14 & 580.76$\pm$2.51 & 41.61$\pm$0.11 & 584.84$\pm$2.03 & 41.64$\pm$0.21 & 582.74$\pm$1.18 & \textbf{41.71$\pm$0.17} & 583.70$\pm$1.95 & 41.50$\pm$0.27 & 590.90$\pm$1.40 & 41.60$\pm$0.15 & 589.44$\pm$1.23 \\
 & OCL & 43.60$\pm$0.31 & 581.54$\pm$0.52 & 43.73$\pm$0.27 & 581.98$\pm$1.18 & 43.50$\pm$0.26 & \textbf{581.37$\pm$0.40} & \textbf{43.86$\pm$0.11} & 582.63$\pm$1.47 & 43.68$\pm$0.26 & 582.26$\pm$1.07 & 43.74$\pm$0.35 & 589.63$\pm$1.17 & 43.65$\pm$0.41 & 591.88$\pm$3.68 \\
 & SL & 45.16$\pm$0.23 & \textbf{584.80$\pm$0.75} & 44.90$\pm$0.33 & 585.97$\pm$1.66 & 44.75$\pm$0.42 & 588.60$\pm$2.74 & 44.98$\pm$0.14 & 588.49$\pm$2.93 & 44.88$\pm$0.23 & 587.16$\pm$0.77 & \textbf{45.20$\pm$0.18} & 593.36$\pm$0.75 & 44.93$\pm$0.22 & 594.54$\pm$1.03 \\\hline
Yelp  & BATL & 45.74$\pm$0.23 & \textbf{607.66$\pm$3.46} & 45.71$\pm$0.18 & 623.79$\pm$0.50 & 45.97$\pm$0.11 & 625.05$\pm$1.34 & 45.77$\pm$0.09 & 624.96$\pm$1.02 & 45.90$\pm$0.16 & 624.53$\pm$0.88 & 45.82$\pm$0.15 & 619.65$\pm$1.68 & \cellcolor{yellow!75}\textbf{46.19$\pm$0.12} & 614.65$\pm$1.70 \\
(5000) & CTL & 40.88$\pm$0.27 & \cellcolor{green!75}\textbf{590.04$\pm$2.02} & 40.93$\pm$0.23 & 594.08$\pm$1.76 & 40.92$\pm$0.02 & 594.33$\pm$0.79 & 40.98$\pm$0.23 & 594.59$\pm$1.29 & \textbf{41.04$\pm$0.27} & 596.56$\pm$1.70 & 40.84$\pm$0.20 & 604.23$\pm$2.38 & 40.98$\pm$0.37 & 603.21$\pm$1.27 \\
 & OCL & 43.91$\pm$0.24 & \textbf{604.26$\pm$1.40} & 43.87$\pm$0.24 & 611.76$\pm$11.11 & 43.79$\pm$0.32 & 607.45$\pm$1.89 & \textbf{44.00$\pm$0.26} & 608.11$\pm$2.32 & 43.71$\pm$0.32 & 606.64$\pm$0.69 & 43.96$\pm$0.20 & 614.86$\pm$2.68 & 43.94$\pm$0.32 & 613.43$\pm$1.70 \\
 & SL & 45.71$\pm$0.12 & \textbf{607.34$\pm$1.31} & 45.50$\pm$0.15 & 608.84$\pm$1.61 & 45.51$\pm$0.22 & 612.08$\pm$0.34 & 45.53$\pm$0.22 & 610.97$\pm$1.14 & 45.32$\pm$0.25 & 611.40$\pm$0.81 & 45.75$\pm$0.20 & 618.60$\pm$1.74 & \textbf{45.76$\pm$0.13} & 617.86$\pm$1.26 \\\hline

\end{tabular}
}
\end{table}

\end{landscape}

In future works, we intend to (a) extend to different stream mining tasks, (b) identify proper moments to automatically trigger the fine-tuning\review{, (c) evaluate the balance between the sampling method and the resources consumption, and (d) evaluate other pre-trained BERT-based models with the combinations of loss functions and sampling method used in this paper}.

\section*{Acknowledgements}

Cristiano Mesquita Garcia is a grantee of a \textit{Doutorado Sanduíche} scholarship provided by \textit{Fundação Coordenação de Aperfeiçoamento de Pessoal de Nível Superior} (CAPES).
This paper is also a byproduct of a CNPq project coordinated by Prof. Jean Paul Barddal under grant number 409371/2022-0 and is tied to CNPq International \#441610/2023-4.

%
%
%
\bibliography{references}
\bibliographystyle{splncs04}

\end{document}